# MULTI MODAL FACE RECOGNITION USING BLOCK BASED CURVELET FEATURES

Jyothi. K [1], Prabhakar C.J [2]

[1] Department of IS&E, J.N.N College of Engineering, Shimoga, Karnataka, India

[2]Department of Studies and Research in Computer Science *Kuvempu University, Shankaraghatta, Karnataka, India*

*ABSTRACT*

*In this paper, we present multimodal 2D +3D face recognition method using block based curvelet features. The 3D surface of face (Depth Map) is computed from the stereo face images using stereo vision technique. The statistical measures such as mean, standard deviation, variance and entropy are extracted from each block of curvelet subband for both depth and intensity images independently.In order to compute the decision score, the KNN classifier is employed independently for both intensity and depth map. Further, computed decision scoresof intensity and depth map are combined at decision level to improve the face recognition rate. The combination of intensity and depth map is verified experimentally using benchmark face database. The experimental results show that the proposed multimodal method is better than individual modality.*

*Keywords*

*Curvelet Transform, Multimodal face recognition, Depth, Disparity, Stereo,  Block- based.*

## 1. INTRODUCTION

Face recognition is of more importance in many applications such as human-machine interfaces, personal identification, image retrieval, and employee access to high security areas. Face recognition algorithms are categorized into 1) 2D face recognition, 2) 3D face recognition and 3) multimodal face recognition algorithms (2D and 3D facial data). Most of the research in face recognition has focused on 2D images. The 2D face recognition methods are sensitive to the changes in pose, illumination, and face expression. A robust identification system may require the fusion of several modalities because ambiguities in face recognition can be reduced with multiple-modal information fusion. A multimodal recognition system normally performs better than any one of its individual components. One of the multimodal approaches is 2D plus 3D [1–4]. A good survey on 3D, 3D-plus-2D face recognition can be found in [1]. Bowyer et al. [1] give a survey of 3D and multimodal face recognition algorithms and state that multimodal face recognition performs better than 2D and 3D face recognition alone. The 3D data of the human face naturally provides high discriminatory information and is less sensitive to variations in environmental conditions such as illumination, skin-color, pose, and makeup; that is, it lacks the intrinsic weakness of 2D approaches.

There have been several face recognition methods based on 3D information were introduced [5-6].Huang, Blanz and Heisele [5] reported a 3D face recognition method which uses a morphable 3D head model to synthesize training images under a variety of conditions. Zhang and Samaras [6] used Blanz and Vetter's [7] morphable model for recognition. The method is introduced to perform well in case of multiple illuminants. The mahalanobis distance method is used for classification. Basri and Jacobs [8] proposed a technique, in which a set of images of a convex Lambertian object obtained under arbitrary illumination can be accurately approximated by a 3D linear space which can be analytically characterized using surface spherical harmonics. Ansari and Abdel-Mottaleb [9] introduced a method based on a stereo images landmark points around the eyes, nose and mouth were extracted from the 2D images and converted to 3D landmark points. They used the CANDIDE-3 model [10] for face recognition. A 3D model was obtained by transforming the CANDIDE-3 generic face to match the landmark points. The eyes, mouth and nose of the 3D model were independently matched for the face recognition and achieved a recognition rate of 96.2% using a database of 26 subjects. The quality of 3D facial data is not good as that of 2D images; because of missing data.But 2D images provide detailed information of face with lack of depth information. The lack of depth information in 2D image can solve by combining 2D and 3D facial information, which can improve the face recognition performance.

The multimodal 2D+3D face recognition can be divided into two methods: early stage and later stage fusion approaches. Early stage method includes pixel, image, and data level fusions. The later stage approaches include feature, score and decision level fusion. The later stage is the most popular fusion technique.

There are three main methods to capture 3D facial information [11]: use of laser range system, structure light method and by stereo scopic technique. The images captured by using laser range system are more reliable and robust to change of color and illumination. Xue and Ding [12] proposed a multimodal boosting algorithm in which 3D range data is fused with 2D face data. Depth and texture maps are combined for identification and verification. A. Leonardis et al. [13] proposed a hybrid (feature-based and holistic) matching for the 3D faces and a holistic matching approach on the 2D faces. Tsalakanidou et al. [14] proposed that exploit depth information to achieve robust face detection and localization under conditions of background clutter, occlusion, face pose alteration, and harsh illumination. The embedded hidden Markov model technique is applied to depth maps and color images. Beumier et al. [15] proposed an automatic face recognition algorithm using face profile data of 2D and 3D facial images. The full facial surface is constructed based on geometric features of the external contour along with the profile-based approach. Both frontal and profile view of 2D face data for identification through the combination of face classifiers is reported in [16]. While the profile view alone provides lower performance than combination of frontal view and profile view.

Lu et al. [17] used feature detection and registration with the ICP algorithm in the 3D domain and LDA in the 2D domain for multimodal face recognition. They handle pose variations by matching partial scans of the face to complete face models. The 3D faces are also used to synthesize novel appearances (2D faces) with pose and illumination variations in order to achieve robustness during 2D face matching. They reported a multimodal recognition rate of 99 percent for neutral faces and 77 percent recognition rate for smiling faces for database of 200 gallery and 598 probe faces. Kusuma and Chua [18] reported PCA based 2D + 3D face recognition method, which uses texture and 3D range facial information. Bronstein et al. [19] presented an expression-invariant multimodal face recognition algorithm. They assume that 3D facial surface to be isometric and remove the effects of expressions by finding an isometricinvariant representation of the face. They used a database with only 30 subjects and did not discuss how an open mouth expression is handled by their algorithm. Tsalakanidou et al. [20] reported the use of depth information to improve face recognition. The recognition algorithm uses Eigen value of color and range image. They have evaluated three different approaches like color, depth and combination of color and depth for face recognition and quantify the contribution of depth. They used VRML models of the XM2VTS database for depth information.

The cost of laser range system is more and which is not desirable in many application.The most cost-effective solution is to use two or more digital cameras to acquire images simultaneously; this method is called stereo acquisition.The advantages of stereo acquisition system is that the acquisition is fast and the distance to the camera scan be adjusted. L.E.Aik et al. [21] proposed a passive stereo vision based face recognition system. The disparity map is obtained using correlation based sum of absolute difference method. To reduce the dimension Fuzzy C-Means clustering is applied on disparity vector. Then PCA transformation is estimated for an intensity image and disparity map separately. The mahalonobis distance technique is used to find the distance between features. R. Fransens [22] presented an algorithm to builds a stereo reconstruction of the face by adjusting the global transformation parameters and shape parameters of a 3D morphable face model. The resulting shape and texture information from a face is used for face recognition. To evaluate algorithm, an extensive stereo image database was built.

Wang et al. [11] proposed a face recognition method by combining the appearance and disparity obtained by passive stereoscopy. The facial features of the appearance and disparity image are extracted by using bilateral two dimensional linear discriminant analysis. Lao et al. [23] perform 3D face recognition using a sparse depth map obtained from stereo images. Combination of 2D edges and iso-luminance contours are used in finding the irises. The face recognition is performed by the closest average difference in corresponding points after the data are converted to a canonical pose. Yin and Yourst [24] exploited reconstructing the 3D shape from front and profile images of the person using a dynamic mesh. A curvature-based descriptor is computed for each vertex of the mesh. Shape and texture features are used for matching operation. T H Sun and FangChin [25] introduced a face recognition that combines 2D and disparity face. The features extracted with principal component analysis and local autocorrelation coefficient (LAC) respectively. The disparity map is obtained by using a synchronous Hopfield neural network stereo matching method. The face features was trained and classified using back propagation neural network technique. T.H.Sun and M.Chen [26] proposed a face recognition technique based on the stereo matching, and Eigen face technique. The 3D facial data was obtained by asynchronous Hopfield neural network stereo matching algorithm. The smallest mahalanobis distances are used for recognition.

The face recognition system consists of two subtasks: feature extraction and classification. Feature extraction is a main step in face recognition. A good feature extraction technique can enhance the performance of any face recognition system. Transformation based approaches have been proposed to improve the performance of a face recognition system for images with high dimensionality. Development of enhanced multi resolution analysis techniques has encouraged research community to apply these tools to achieve a high level of performance in pattern recognition applications. The several problems such as, deformation of face images, illumination and contrast variations can be solved by extracting features using multiresolution technique. Xu et al.[27] proposed multimodal recognition method. The features of 2D and 3D information are extracted by using Gabor filters and Linear Discriminant Analysis (LDA) is used for dimensionality reduction. Wang et.al.[28]used Gabor filter features in 2D images and point signatures in the 3D range data to perform multi modal face recognition. The 2D and 3D features are combined at feature level. Support vector machines with a decision directed acyclic graph is used for classification. Kautkar et.al [29] developed a method that combines the fisherface with ridgelet transform and high speed photometric stereo.

The curvelet transform [30] is one of the most popular multiresolution techniques. The curvelet transform produces a more sparse representation of the image than wavelet and ridgelet transform. Emergence of curvelets [30] that offer enhanced directional and edge representation. So far, it is used in the field of image denoising, image compression, image enhancement, etc. but not much work has been done in curvelet based pattern recognition. The curvelet based 2D face recognition has been presented in [31-34]. In [31], the feature extraction has been done by decomposing the 2D original image using curvelet transforms and its quantized 4 bit and 2 bit representations.Three Support Vector Machines are used for classification of three sets of coefficients. The results obtained by three SVMs classifier are fused to determine the final classification. The experiments were carried out on three well known databases; those are the Georgia Tech Face Database, AT&T "The Database of Faces" and the Essex Grimace Face Database.The work presented in [32] is an extension of [31]. In [32], used the bit quantized images to extract curvelet features at five different resolutions. The 15 sets of approximate coefficients are used to train 15 Support Vector Machines and results are combined using majority voting. In [34] is proposed aface recognition algorithm to reduce the computation cost of [33]. To reduce the dimension of curvelet features, dimensionality reduction tools like PCA and LDA method and evaluated by conducting different experiments on three different databases: ORL, Essex Grimace and Yale Face Database.

In this paper, we proposed a multimodal 2D +3D face recognition technique.  3D data or depth map of face images is extracted from stereo face images using adaptive weight based stereo matching method. The multiresolution technique such as curvelet transform is applied independently on depth and intensity image of faces which decompose the face image into subbands at various scales. The Figure 2 shows three levels of curvelet decomposition applied on face image. For each level, the curvelet transform decompose the image into its approximate and detailed subbands. These subbands (subimages) obtained are called curvefaces. These curvefaces greatly reduce the dimensionality of the original image.In order to avoid the redundancy of curvelet coefficients in each subband, we divide the curvelet subbands into equally sized blocks.We extract compact and meaningful feature vectors using simple statistical measuressuch as Mean, Variance, Entropy and Standard Deviation from each block of curvelet subband. The classifier is used to measure decision scores of 2D and 3D facial images independently. The estimated decision scores of 2D and 3D features are fused at the decision level using OR rule. The proposed method is evaluated by carrying out experiments on the stereo face database [22].

The remaining sections of the paper are organized as follows. The adaptive support weight based stereo matching method is discussed in section 2. In section 3, a brief description of curvelet transforms is presented. The block based curvelet feature extraction technique is explained in section 4. In section.5, the fusion of decision scores of intensity and depth map of image is discussed .The experimental results are presented in section 6. Finally, the conclusion is given.

## 2. STEREO DEPTH MAP OF FACE

The disparity map is estimated by considering sum of square difference of intensity and gradient of each pixel of stereo pairs. The dissimilarity measure based on the gradient fields and intensities of the images yields good accuracy in the resulting disparity map. The main reason for this accuracy enhancement is the better robustness of the gradient against differences in sampling and local brightness changes between cameras. That is defined as

$$SSD(q, q_d) = \sum_{q,q_d \in W(x,y)} (b_1(q) - b_2(q_d))^2, \qquad (4)$$

$$Grad(q, q_d) = \sum_{q,q_d \in W_x(x,y)} |\nabla_x b_1(q) - \nabla_x b_2(q_d)|^2 + \sum_{q,q_d \in W_y(x,y)} |\nabla_y b_1(q) - \nabla_y b_2(q_d)|^2, \qquad (5)$$

where $W$ is 11×11 window. The left to right cross checking is used for occlusion handling and to find reliable correspondences. The resulting dissimilarity measure is given by

$$C(q, q_d) = SSD(q, q_d) + Grad(q, q_d). \qquad (6)$$

After the dissimilarity computation, winner takes all optimization is used to choose the disparity with lowest matching cost. In our method, instead of assigning disparity value to each pixel, the disparity plane is estimated by using disparities of each segment. The estimated disparity plane is assigned to each segment using least square plane fitting method.After estimating the disparity face; the depth map is calculated using Delaunay triangulation technique:

$$D = b*f/d, \qquad (8)$$

Where $D$ represents the depth, $d$ is the disparity, $b$ is the baseline and $f$ is the focal length.

## 3. CURVELET TRANSFORM

Candes and Donoho [30] proposed a new multi-resolution analysis tool such as curvelet transforms. Curvelet transform is multiscale and multidirectional transform. Curvelets exhibit highly anisotropic shape obeying parabolic scaling relationship. Curvelets are elongated needle shaped structures. Curvelet transform can efficiently represent edge discontinuities in images. The coefficients of curvelet are more sensitive to orientation, which is a useful property for detecting curves in images. The curvelet transform produces smooth image except for discontinuity along a general curve. Generally, in order to extract curvelet features, the images are decomposed into its approximate and detailed components for various levels. These sub images greatly reduce the dimensionality of the original image.

In curvelet transform, 2D fast Fourier transform (FFT) of the image is taken. Then this image is divided into parabolic wedges. Finally, an inverse FFT of each wedge is used to find the curvelet coefficients at each scale $j$ and angle $l$. The wedges are gotten from partitioning the Fourier plane in radial and angular divisions. Each wedge corresponds to a particular curvelet at that given scale and angle. There are two types of digital curvelet transform, they are

- Curve-let via unequally spaced FFT
- Curve-let via wrapping. In our work we used FDCT Wrapping technique.

The discrete curve-let transform can be defined by a pair of windows, such as a radial window $W$ and an angular window $V$.

$$W_j(\omega) = \sqrt{\varphi_{j+1}^2(\omega) - \varphi_j^2(\omega)}, \; j \geq 0, \tag{9}$$

Where $\varphi$ is a product of low pass one dimensional windows

$$\varphi_j(\omega_1, \omega_2) = \emptyset(2^{-j}\omega_1)\emptyset(2^{-j}\omega_2), \qquad 0 \leq \emptyset \leq 1. \tag{10}$$

$$V_j(\omega) = V(2^{\frac{j}{2}}\omega_1/\omega_2). \tag{11}$$

The Cartesian window $U_j$ isolates the frequencies near the wedge $(\omega_1, \omega_2)$ is defined as

$$U_{j,l}(\omega) = W_j(\omega)V_j(\omega) \tag{12}$$

Then curvelet transform can be represented as

$$c(j, l, k) = (f, U_j) \tag{13}$$

$j, l \; and \; k$ is scale, orientation and position variable respectively.

The discrete curvelet via wrapping algorithm is as follows:

1. Obtain Fourier samples using 2D FFT $f[n_1, n_2], -n/2 \leq n_1, \; n_2 < n/2$.
2. For each scale $j$ and angle $l$ form the product $U_{j,l}[n_1, n_2]f[n_1, n_2]$.
3. Wrap this product around the origin.
4. Estimate the discrete coefficients $c(j, l, k)$ by applying inverse 2D FFT.

## 4. BLOCK BASED CURVELET FEATURES

Curvelet transform is applied to intensity and stereo depth map of faces independently to decompose into various subbands. The dimensionality of curvelet features is extremely large in size, because of multiple scale and orientations. For instance, the size of the depth image or intensity image is $64 \times 64$ and the size of curvelet feature vector is 30833 for scale=5 and orientation=8. Thus, the dimensionality of curvelet feature vector is very high in each subband due to redundant and irrelevant information. For an efficient representation of the face image, first each subband is divided into an equally sized blocks of $8 \times 8$. If the dimensions of the image are not integral multiples of 8, pads zeroes along the rows and/or columns to get the integral number of blocks. These local features are more robust against variations in pose or illumination. Then statistical measures such as mean, variance, standard deviation and entropy are applied to each block of the subband. The statistical measures of blocks are then concatenated to form a global description of the face. This reduces the dimensionality of the curvelet feature vector from 30833 to 2024.

The three reasons for selecting the block size to be $8 \times 8$ are [39]:

- It is an adequate size in order to collect the information from all the regions of the face.
- Useful image contents vary gracefully within a block.
- Size of the block is best suited for meeting both the processing and timing constraints of most of the real time processors.

Let $I_{ij}(p,q)$ be the image at the specific block $j$ of subband $i$, the resulting feature vector

$$C_{ij} = \{\mu_{ij}, \sigma_{ij}, \sigma_{ij}^2, e_{ij}\}, \tag{14}$$

where $\mu_{ij}, \sigma_{ij}, \sigma_{ij}^2, e_{ij}$ are mean, standard deviation, variance and entropy respectively, which are extracted for each block and are defined as:

$$\mu_{ij} = \frac{1}{M \times N} \sum_{p=1}^{M} \sum_{q=1}^{N} I_{ij}(p,q), \tag{15}$$

$$\sigma_{ij} = \frac{1}{M \times N} \sum_{p=1}^{M} \sum_{q=1}^{N} [I_{ij}(p,q) - \mu_{ij}]^2, \tag{16}$$

$$\sigma_{ij}^2 = \left\{ \frac{1}{M \times N} \sum_{p=1}^{M} \sum_{q=1}^{N} [I_{ij}(p,q) - \mu_{ij}]^2 \right\}^{1/2}, \tag{17}$$

$$e_{ij} = -\sum_p (h \times \log(h)), \tag{18}$$

Where $M$ and $N$ is the size of $I_{ij}(p,q)$ and $h$ is the histogram counts.

The feature vector of a face is then obtained by concatenating each feature vectors of sub-bands $C = \cup_{i=1}^{m} \cup_{j=1}^{n} \{C_{ij}\}$, $m$ is the number of subbands and $n$ is the number of blocks in each subband. The feature vectors of intensity and depth map are extracted and stored independently.

## 5. FUSION OF INTENSITY AND DEPTH MAP

The performance of the multi modal face recognition system depends on matter of how to fuse two or more sources of information. The high recognition rate can be achieved, if there is low mutual correlation among individual decisions. The intensity and depth information are highly uncorrelated, because depth data gives surface of face and intensity information yields texture of the surface. The fusion of two or more information can be done at the feature level, score level, or at decision level. Feature level fusion fuses multiple extracted features from each biometric. Score level fusion combines the matching scores from each biometric. Decision level fusion uses the results from different classifiers and makes the final decision based on all of them. In this paper, we are interested in the fusion at the decision level. In the first step, classifier is used to estimate the decision scores for each set of block based curvelet facial features of both intensity and depth map independently. Each query image is examined based on the minimum distance of its features from the features of other images in the training samples. In the second step, we combine decision scores obtained for each set of extracted curvelet features of both intensity and depth map. We employ OR rule inorder to combine decision scores obtained for each set of curvelet features. With two decision scores yielded by classifier for two sets of curvelet features, the fusion rule can be represented logically as illustrated below:

| X | Y | F |
|---|---|---|
| 0 | 0 | $D_0$ |
| 0 | 1 | $D_1$ |
| 1 | 0 | $D_2$ |
| 1 | 1 | $D_3$ |

In this table, *X* is the decision output by the first classifier and *Y* is the decision output by second classifier. The OR fusion rule is represented as

$$F = X \mid Y \qquad (19)$$

This rule replaces $\{D_0, D_1, D_2, \text{and } D3\}$ with $\{0, 1, 1, \text{and } 1\}$.

## 6. EXPERIMENTAL RESULTS

To evaluate the performance of the proposed multi modal face recognition method, we carried out various experiments using stereo face database [22]. It consists of stereo pairs of 70 subjects (35 males, 35 females) recorded from eight different viewpoints. We have considered subset of the database which consists of 50 subjects. The facial portion is extracted based on nose tip as reference andthe noises in the cropped images are filtered by using median filters. The cropped color images are changed to the gray-level image by averaging R, G and B color components. The cropped color images of three subjects with eight different views and corresponding estimated stereo depth map are shown in Figure 1. In order to obtain the optimal model, 4-fold cross validation is performed on the training set. The 4-fold cross validation method divides the training set into 4 subsets of equal size. For each one of 4 experiments, 3 folds are used for training and the remaining one is used for testing. These results have been achieved by averaging the recognition rates of four different rounds of face recognition. The advantage of cross validation is that all the images in the dataset are used for both training and testing.K nearest neighbor (KNN) classifier is used for classification, which uses K closest samples to the query image. The performance of KNN classifier depends on the value of the K [40].

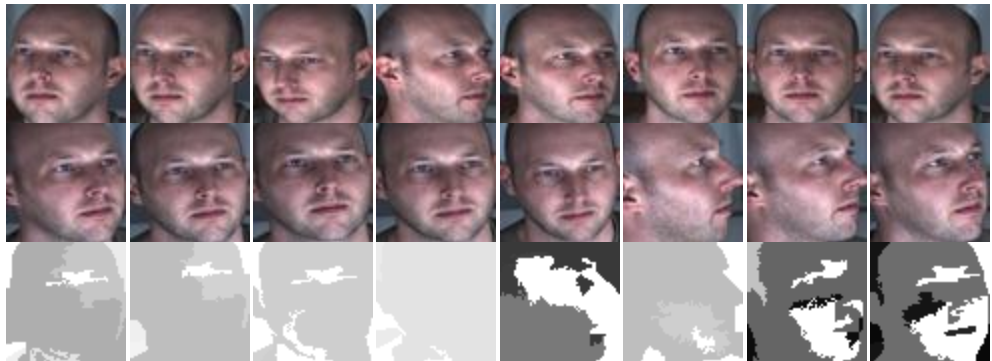

(a)

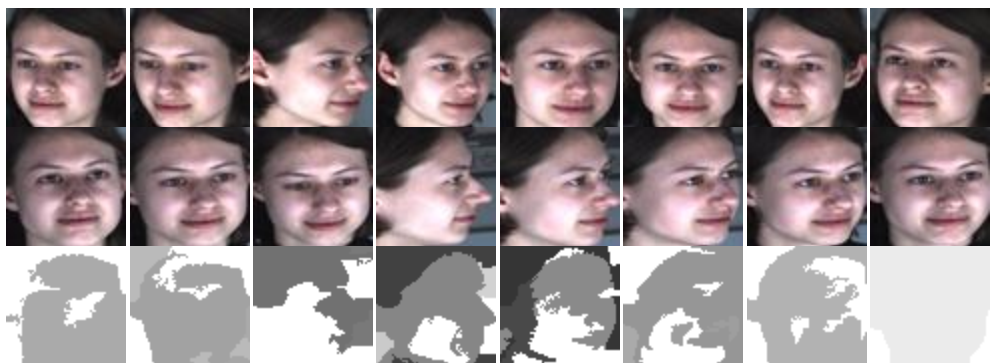

(b)

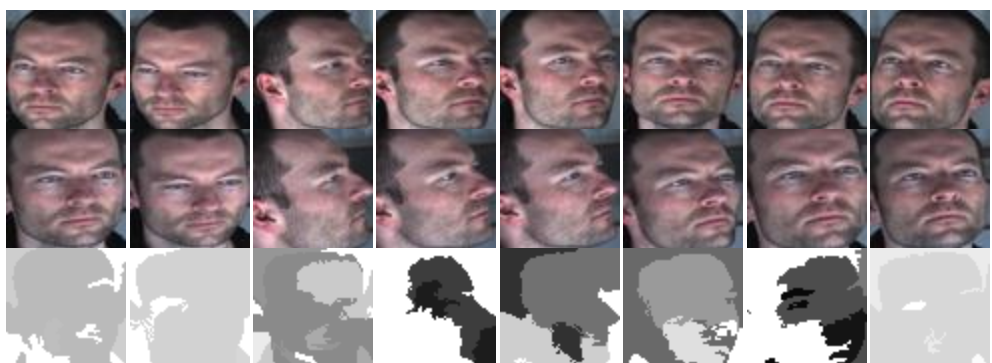

(c)

Figure 1.Three sets of images (a),(b),(c) represents three subjects of stereo database; In each set, first and second row shows left and right face image ; third row shows corresponding estimated depth map.

We considered the left image as the intensity image and depth image is obtained using pair of training images of same person by employing stereo vision technique explained in the section 2. The curvelet transform is applied on depth and intensity face image to decompose into subbands. Figure 2 shows a result of curvelet transform at scale 3 applied on intensity image of face. The obtained multi orientation information describes face image by a subset of filtered images containing curvelet coefficients which represents the face textures.

The training face image which is divided into blocks of $8 \times 8$ and the first block coefficients with statistical measures are shown in Table 1.

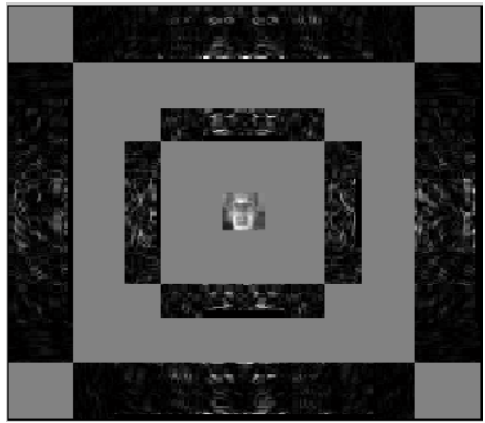

Figure 2.Three levels of curvelet decomposition applied on face image.

Table 1.The curvelet coefficients with statistical measures for the first block of the approximation subband.

| Approximation subband with equal sized Blocks | Coefficients of first block | Mean | Variance | Standard Deviation | Entropy |
|---|---|---|---|---|---|
| 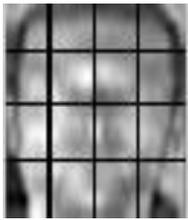 | 17 23 51 72 77 82 91 99<br>16 17 19 30 56 81 94 99<br>16 16 16 16 23 46 80 101<br>20 17 17 16 16 18 33 68<br>25 23 20 17 16 16 17 25<br>35 32 27 22 18 17 16 14<br>53 46 39 30 22 19 18 16<br>67 60 52 41 31 24 21 18 | 36.56 | 683.52 | 26.14 | 0 |
| 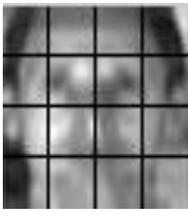 | 104 107 107 107 106 102 95 91<br>104 108 108 108 107 102 94 90<br>106 108 108 108 108 102 93 91<br>100 110 108 107 106 101 93 90<br>56 95 112 106 105 101 91 89<br>21 54 98 112 105 101 92 88<br>14 23 59 102 109 101 91 87<br>16 15 27 65 103 102 91 87 | 90.58 | 688.86 | 26.25 | 0.03 |

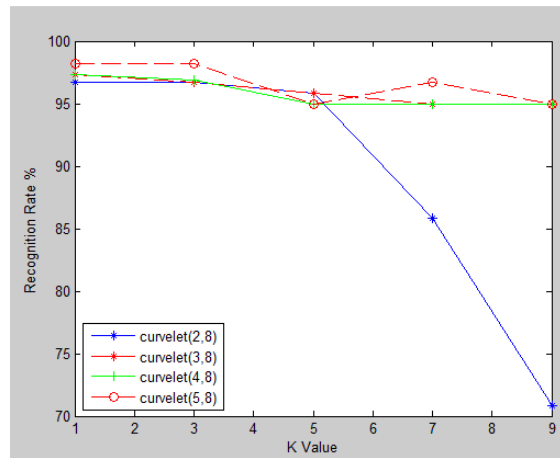

Figure 3. Recognition rates at various K value for different curvelet scales.

We conducted several experiments by varying K value from 1 to 9 and curvelet scale from 2 to 5. The obtained results of experiments allow us to identify, which K value and curvelet scale maximize the recognition accuracy. The plots in Figure 3 shows that the maximum recognition rate can be achieved at K=3 and scale=5, orientation=8. In our all experiment, the images are decomposed using curvelet transform at scale = 5 and orientation =8.The experiments have proved that this scale can balance between recognition rate and recognition time, and larger scales of the transformation will not significantly improve recognition rate.

## 6.1 Single Modal Face Recognition

Chang et.al [41] compared the recognition rate of intensity and depth image independently and concluded that the recognition performance was similar. Here, we evaluated the recognition performance of intensity and depth image obtained by applying adaptive support weight based stereo matching technique on stereo face database [22] independently and draw a different conclusion.

In the first experiment, the PCA algorithm [42] is used for reducing dimensionality of intensity and depth images. In the second experiment, block based curvelet features of intensity image and depth image to describe an individual. The rank one recognition rate of first experiment is shown in the first two columns and of second experiment is shown in the last two columns of Table 3. The block based curvelet features outperform compared to PCA features for intensity images. Similarly, the block based curvelet features of depth map yields highest recognition rate than PCA features for depth map. Hence, we can conclude that block based curvelet features characterize the individual faces thoroughly for face recognition than PCA features for both intensity and depth images. From the above two set of experiments, we can conclude that, block based curvelet features of depth images perform better to characterize the faces for face recognition.

31

Table 2.Comparison of Rank 1 recognition rate for stereo face database.

| Methods | Intensity Eigenface | Depth Eigenface | Intensity block based Curvelet | Depth block based Curvelet |
|---|---|---|---|---|
| Recognition Rate (%) | 83.46 | 84.67 | 85.83 | 87.63 |

## 6.2. Multi Modal Face Recognition

There are three ways to combine the features with different properties. They are feature level, score level, and decision level. The decision level fusion technique is here we used to combine block based curvelet features of depth and block based curvelet features of intensity information. Chang et.al [41] fused the depth and intensity information at score level. But in our first experiment, we applied the PCA to both depth and intensity information independently and decisions obtained are fused at decision level. In the second experiment, the statistical measures such as mean, variance, standard deviation and entropy are estimated from each block of a curvelet subband of depth and intensity images separately.The feature vector of face is obtained by concatenating these statistical measures of each block of curvelet subbands. In both the experiments, the KNN classifier is applied to obtain a class label (decision score). This class labels are fused at decision level using OR rule.

Table 3. Comparison of recognition rate of proposed method for stereo face database

| Methods | Depth +Intensity | Proposed Method |
|---|---|---|
| Recognition Rate (%) | 91.52 | 98.18 |

First column of Table 3 shows rank-one recognition rate for fusion of Eigen features extracted from depth map and intensity information. The second column of Table 3 shows the rank-one recognition rate for fusion of depth and intensity block based curvelet features. It is observed that the fusion of block based curvelet features out performs the fusion of eigenfaces. Also, it is observed that the fusion of features improves the recognition accuracy considerably compared to features of depth and intensity images alone.

Table 4. Comparison of recognition rate of multi resolution methods

| Methods (Depth +Intensity) | Recognition Rate (%) | Computation Time (sec) |
|---|---|---|
| Wavelet | 89.52 | 1.27 |
| Ridgelet | 91.22 | 1.36 |
| Curvelet | 92.13 | 1.42 |
| Proposed method | 98.18 | 1.3 |

In this experiment, the recognition rate and computation times of the various multi resolution recognition methods are compared. The computation time to reconstruct and recognize a pair of stereo test image is calculated for different methods.These experiments were carried out using a PC with Intel core i3 processor with speed 2.27 GHz and 4 GB RAM. The proposed algorithm is implemented using MATLAB R2010a. The recognition performance of various methods is also computed and both are shown in Table 4.

From the results of Table 4, it can be known that the curvelet based method significantly increase the recognition performance in comparison to wavelet based method and ridgelet based method. The ridgelet transform can able to capture multidirectional features, but wavelet transform focus mainly on horizontal, vertical, and diagonal features, which are not dominant in most face images. Curvelet based method had higher recognition rate in comparison to both the wavelet and ridgelet due to the sparse representation of images and extracting crucial edge-based features from facial images more efficiently.So that features obtained from the Curvelet subbands will have more powerful information compared to the features from the wavelet sub-bands and ridgelet subband.

Curvelet based features yielded accuracy rate 92.13%, and proposed method yield recognition rate 98.18% which significantly improved accuracy ranges for ridgelet based features and wavelet based features.The recognition rate for wavelet based method is 89.52%, while ridgelet based method is 91.22 %, and curvelet based accuracy rates were in the 92.13% (without block) and 98.18% (with block). This was also expected since the curvelet transform is able to capture multidirectional features in wedges, as opposed to lines or points as in the ridgelet or wavelet transform.

Table 5.Comparison of Recognition Rate of different classifiers.

| Method | Recognition Rate (%) |
| --- | --- |
| KNN L1 Norm | 92.13 |
| KNN L2 Norm | 98.18 |
| KNN L3 Norm | 94.2 |
| SVM | 74.3 |

We have conducted experiment to compare the accuracy of different classifier and obtained results are shown in Table 5. The KNN classifier is best method for classifying faces based on their images due to its lesser execution time and better accuracy than other commonly used method like Support Vector Machine (SVM). KNN classifier is dominant than SVM in sparse datasets [38]. In practice, training an SVM on the entire data set is slow and the extension of SVM to multiple classes is not as natural as KNN. The recognition accuracy of KNN L2 norm classifier is better than KNN L1, L3 norms and SVM classifier.

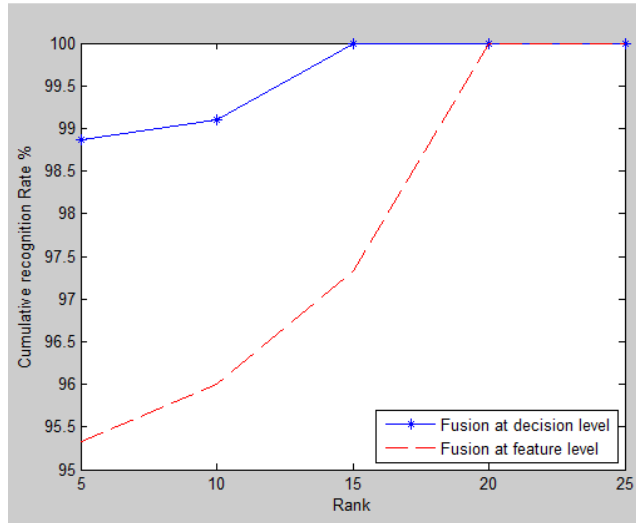

Figure 4. Comparison of two fusion techniques.

Figure 4 shows the cumulative match score (CMS) curves for fusion at the decision level and fusion at feature level. The fusion at the decision level is more robust than the fusion at feature level.

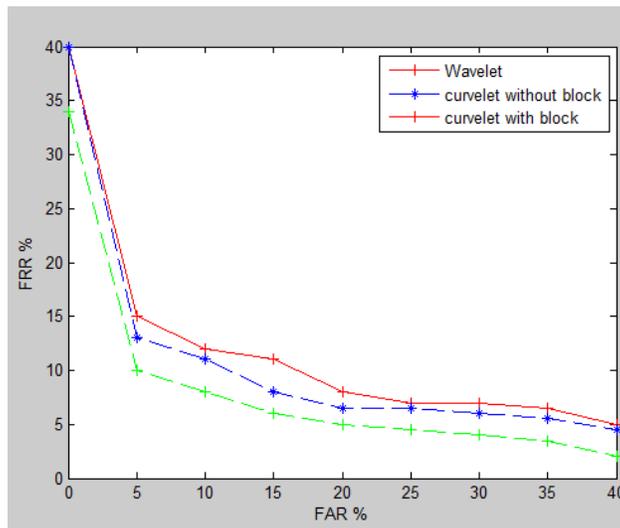

Figure 5. The ROC curves of three different feature extraction method.

The performance of proposed methodology has been evaluated using Receiver Operating Characteristics (ROC) and is shown in Figure 5. It consists of performance measure of False Acceptance Rate (FAR) and False Rejection Rate (FRR) at different values of thresholds. The right recognition rate, is obtained once two feature vectors of the same faces are compared and an imposter matching score is obtained when feature vectors of different faces are compared. FAR is the probability of accepting an imposter as a genuine subject and FRR is the probability of a rejecting a genuine subject. For an individual, images excluding his own images will be the imposters. FRR is computed using following expression.

FRR = (True claims rejected/Total true claims) x 100%  (20)

FAR is computed by counting the number of imposter's claims accepted out of total imposter's claims for given threshold.

FAR = (Imposter claim accepted/Total imposter claims) x 100%  (21)

## 7. CONCLUSIONS

This paper proposes a new approach for face recognition based on exploiting the features of the block based curvelet transform. The depth map of a stereo pair is computed using adaptive support weight based stereo matching technique. The extracted curvelet features of intensity and depth map are extremely large in dimension, since multiple scales and orientations are adopted. To reduce the large dimensions of curvelet features, statistical measures such as mean, variance, standard deviation and entropy are used for selecting effective features of blocks of each subband. These features of eight blocks of all subbands are concatenated to obtain a feature vector of a face. The KNN classifier is employed independently for both intensity and depth map to compute the decision score. These computed decision scores of intensity and depth map are fused at decision level using OR rule.The depth information is more robust than intensity information under pose variations. Thus, the combination of intensity and depth can improve the recognition rate. The results based on fusion of block based curvelet features of depth and curvelet features of intensity image are better than the results based on the fusion of depth eigenface and intensity eigenface.